\documentclass{scrartcl}
\usepackage[utf8]{inputenc}
\usepackage[affil-it]{authblk}
\usepackage[natbibapa]{apacite}
\usepackage{graphicx}
\usepackage{framed}
\usepackage{caption}
\usepackage{newfloat}
\usepackage[hidelinks]{hyperref}

\graphicspath{ {./figures/} }

\DeclareCaptionType{Box}

\newcommand\blfootnote[1]{%
  \begingroup
  \renewcommand\thefootnote{}\footnote{#1}%
  \addtocounter{footnote}{-1}%
  \endgroup
}

\setkomafont{disposition}{\normalcolor\bfseries}

\title{Distributive Justice and Fairness Metrics in Automated Decision-making: How Much Overlap Is There?}

\author[1]{Matthias Kuppler}
\author[1]{Christoph Kern}
\author[1]{Ruben L. Bach}
\author[2,3]{Frauke Kreuter}

\affil[1]{\footnotesize School of Social Sciences, University of Mannheim, Germany}
\affil[2]{\footnotesize Department of Statistics, LMU Munich, Germany}
\affil[3]{\footnotesize Joint Program in Survey Methodology, University of Maryland, USA}

\date{}

\begin{document}

\maketitle

\begin{abstract}
The advent of powerful prediction algorithms led to increased automation of high-stake decisions regarding the allocation of scarce resources such as government spending and welfare support. This automation bears the risk of perpetuating unwanted discrimination against vulnerable and historically disadvantaged groups. Research on algorithmic discrimination in computer science and other disciplines developed a plethora of fairness metrics to detect and correct discriminatory algorithms. Drawing on robust sociological and philosophical discourse on distributive justice, we identify the limitations and problematic implications of prominent fairness metrics. We show that metrics implementing equality of opportunity only apply when resource allocations are based on deservingness, but fail when allocations should reflect concerns about egalitarianism, sufficiency, and priority. We argue that by cleanly distinguishing between prediction tasks and decision tasks, research on fair machine learning could take better advantage of the rich literature on distributive justice.\blfootnote{\textbf{Acknowledgements:} Christoph Kern's and Ruben Bach's work was financed by the Baden-W{\"u}rttemberg Stiftung (grant “FairADM -- Fairness in Algorithmic Decision Making” to Ruben Bach, Christoph Kern and Frauke Kreuter.}
\end{abstract}

\section{Introduction} \label{section1}
The rise of prediction-based or statistical decision-making \citep{mitchell_prediction-based_2020} renewed interest in the automation of allocation decisions in many domains that impact people's life chances. For instance, public agencies experiment with automating the allocation of support to jobseekers based on predicted unemployment risk \citep{allhutter_algorithmic_2020,niklas_profiling_2015}, the allocation of police forces to neighbourhoods based on predicted risk of burglary \citep{gerstner_predictive_2018}, and the allocation of intervention and supervision resources to convicted offenders based on predicted re-offending risk \citep{howard_construction_2012}. Automation is framed as an instrument to increase the objectivity of allocation decisions by eliminating human biases \citep{lepri_fair_2018} but also as a powerful tool to uncover relations in data that would stay hidden to humans. However, at the same time, concerns are increasingly raised that automation may create a new source of unwanted discrimination against vulnerable and historically disadvantaged groups \citep{barocas_big_2016}.

These concerns sparked a lot of research on formalizing notions of justice into metrics that detect and mitigate discrimination and unfairness in automated decisions in recent years \citep{makhlouf_applicability_2020, mehrabi_survey_2019, mitchell_prediction-based_2020, verma_fairness_2018}. However, so far, the link between algorithmic fairness and distributive justice is poorly understood and often ignored in the \textit{fairML} literature. Consequently, existing work builds upon an overly narrow notion of justice drawn mostly from anti-discrimination legislation rather than from robust philosophical discourse. Setting important differences between fairness metrics aside, prominent metrics consider an allocation decision just if people who only differ on protected attributes (sex, ethnicity, disability etc.) are assigned identical decisions with identical error probability. It is not clear what kind of distributive justice is served by such metrics, in which contexts such metrics are justified, and why certain attributes are protected while others are not. In other words, the question of who should get how much is often ignored in discussions of fairness in algorithmic decision-making.

To address this shortcoming, we systematically trace the link between algorithmic fairness and distributive justice. Distributive justice is concerned with the allocation of goods and burdens among members of a society \citep{lamont_distributive_2017}. Here, we do not address related but somewhat different concepts of retributive \citep{walen_retributive_2020} and procedural justice \citep{miller_justice_2017}. They are less relevant for the kind of automated decision-making discussed in the literature as they address the allocation of punishment (retributive justice) as in criminal law and the fairness of decision processes (procedural justice) as in the literature on transparent machine learning \citep{rudin_stop_2019}. Thus, trading depth for breadth of discussion, we focus on distributive justice alone. 

To set the stage for our discussion, Section \ref{section2} presents the basic setup of prediction-based decision-making. Section \ref{section3} reviews prominent distributive justice theories that we will need to  understand in order to connect distributive justice with algorithmic fairness. Section \ref{section4} reviews existing fairness metrics as discussed in the fairML literature and investigates their connection to distributive justice theory. Section \ref{section5} clarifies the relation between distributive justice and fairness metrics by clearly distinguishing between prediction and allocation.

Note that we are not the first to explore the link between fairness metrics and distributive justice theories. \cite{binns_fairness_2018} provides an excellent primer on distributive justice for computer scientists but does not elaborate on the question how distributive justice theories are translated into the mathematical formalism of fairness metrics. \cite{gajane_formalizing_2018} and \cite{lee_fairness_2020} show the connection between selected fairness metrics and their corresponding concepts in distributive justice. Their reviews, however, lack a discussion of distributive justice theories that have not been formalised in the fairML literature so far.

Therefore, our contribution to the literature is twofold: On the one hand, we provide an accessible discussion of the most prominent distributive justice theories from an automated decision-making perspective. On the other hand, we systematically discuss the link between distributive justice and fairness metrics. Thereby, we highlight which notions of justice are prevalent in the literature and which are neglected. Overall, our goal is to enlarge computer science’s decision space when it comes to formalising fairness metrics and designing allocation systems.

\section{Setup and Notation} \label{section2}
Consider a public agency with a fixed amount of public resources that it can allocate among its clients. The public resource is rivalrous: Once a given amount of the resource is allocated to one client, it is gone and cannot be allocated to another client. The public resource is scarce, meaning that there might not be enough of the resource to satisfy the needs of all clients. The task of the public agency is to find a distribution rule that defines how the public resource should be allocated across clients. We denote this step as the decision task. The distribution rules take the form: Allocate amount \(R\) of the resource to client \(X\) if and only if (iff)  \(X\) has attribute  \(Y\). In automated allocation systems, the distribution rule supports or replaces human decision-makers.

The attribute  \(Y\) might be unobserved at decision time such that the public agency must predict \(Y\) from client \(X\)’s observed attributes \(V\). Observed attributes are separable into protected attributes  \(A\)  (e.g. sex, ethnicity, religion) and unprotected attributes  \(B\). The prediction task is a classification problem for categorical  \(Y\) and a regression problem for continuous  \(Y\). Call \(Y\) the target. Let \(v_i=\{v_{i1},…,v_{ik}\}\) be the feature vector of the \(i\)-th client. The feature vector is a \(k\)-dimensional vector of numerical values that represent the attributes of the client. The prediction task is: Given a training dataset of \(N\) clients of the form \(\{(v_1,y_1 ),…,(v_N,y_N )\}\) such that \(v_i\) is the feature vector of the \(i\)-th client and \(y_i\) is the true target value of the \(i\)-th client, find a function \(g:V \rightarrow Y\) that maps the feature vector onto the target. Let \(g(v_i) = \hat{y}_i\) be the predicted value of client \(i\)’s true target value \(y_i\). 

In prediction-based decision-making, \(\hat{y}_i\) is plugged into the distribution rule to determine the amount \(R\) of the resource that should be allocated to the client. That is, the decision task is informed by the result of the prediction task.

\captionof{Box}{Example of Automated Allocation\label{box:jobs}}
\begin{framed}
As an illustration of automated decision making in the public sector, consider the prediction-based allocation of support programs to job seekers \citep{allhutter_algorithmic_2020,niklas_profiling_2015}. The decision task includes assigning a limited number of support programs to job seekers, e.g., upon their entry into unemployment. As the target attribute, employment agencies may consider the unemployment status six month after registering as unemployed. The prediction task includes inferring that status at decision time by learning the link between, e.g., employment histories, demographic information and subsequent unemployment status from historical data.
\end{framed}

In the next section, we review distributive justice theories in detail.

\section{Distributive Justice} \label{section3}
Distributive justice is concerned with the allocation of goods and burdens among members of a society under conditions of relative material scarcity \citep{lamont_distributive_2017, olsaretti_introduction_2018}. At the core of distributive justice theories are well-justified distribution principles and metrics of justice. Distribution principles are rules that define on what basis goods and burdens should be distributed. The justice metric defines the type of goods or burdens (resources, welfare, capabilities) to which the distribution principle should be applied.

For the sake of exposition, we will focus on the allocation of goods and utilize resources as the justice metric. Resources are things, material and immaterial, that individuals can use to realize valuable states of being and doing (e.g., being well-nourished) from which they derive well-being. We set aside the problem that individuals differ in their ability to transform resources into well-being \citep{robeyns_capability_2020} and assume that a given amount \(R\) of the resource is equally valuable for all individuals.

Contemporary distributive justice theories all start from the “egalitarian plateau” \citep[p. 4]{kymlicka_contemporary_2002} of the moral equality of all humans. All humans have the same fundamental worth and dignity and should be treated with equal concern and respect \citep{arneson_egalitarianism_2013, gosepath_equality_2011}. Distributive justice theories differ in the kind of equal treatment that is derived from moral equality.

The following sections discuss the most prominent distributive justice theories along with their respective distribution principles: Egalitarianism (\ref{section31}), desert (\ref{section32}), sufficiency (\ref{section33}), priority (\ref{section34}), and equality of opportunity (\ref{section35}). Section \ref{section36} discusses the role of luck and responsibility. Table \ref{tab1} summarizes the distribution principles and illustrates possible applications to the example allocation system presented in Box \ref{box:jobs}.

\begin{table}
\caption{Distribution Principles}
\includegraphics[width=\textwidth]{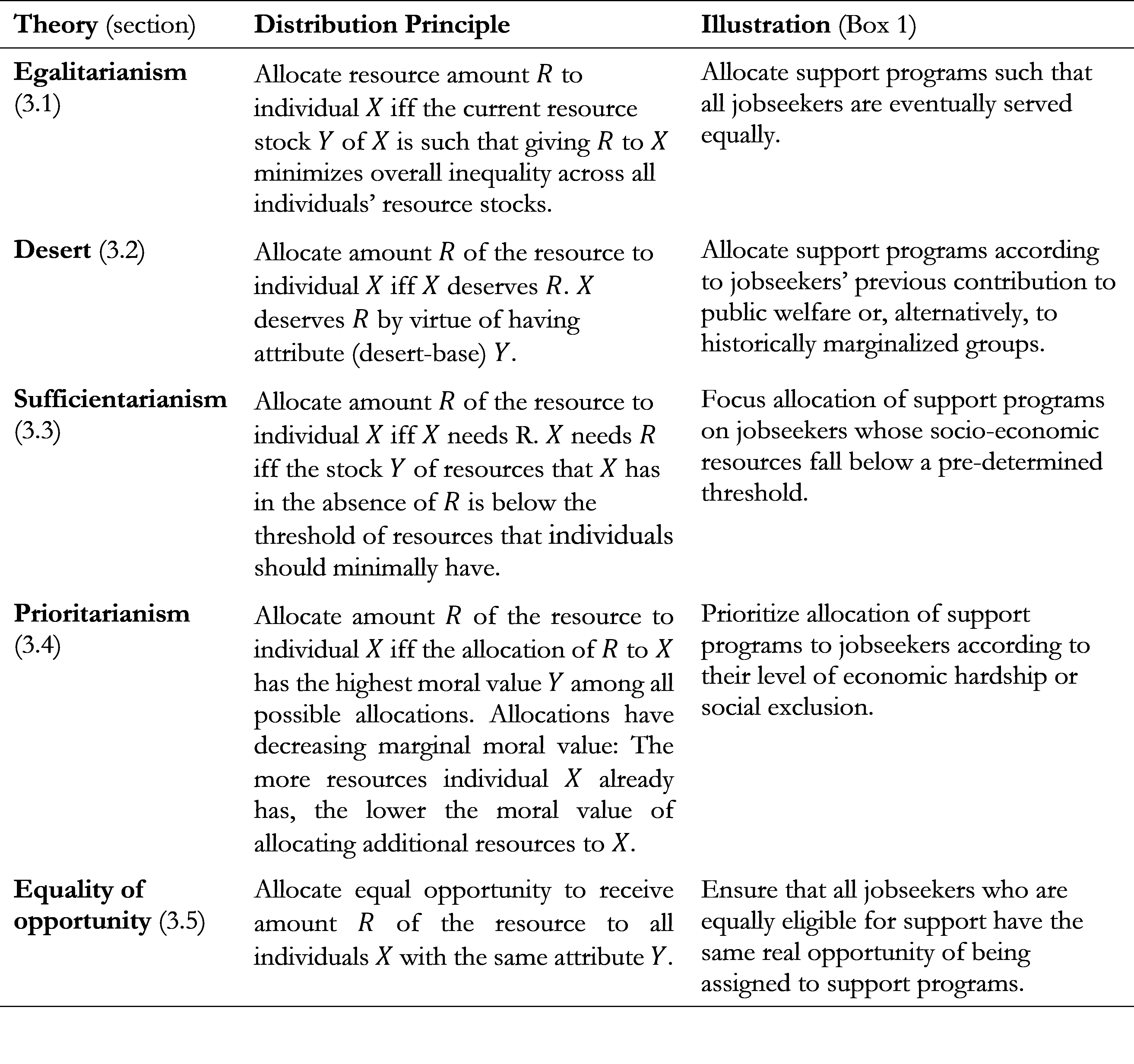}
\centering
\label{tab1}
\end{table}

\subsection{Egalitarianism} \label{section31}
The distribution principle of (strict) egalitarianism requires allocating resources such that overall resource inequality across individuals is minimized \citep{arneson_egalitarianism_2013, gosepath_equality_2011}. More formally: Allocate resource amount \(R\) to individual \(X\) if the current resource stock \(Y\) of \(X\) is such that giving \(R\) to \(X\) minimizes overall inequality across all individuals’ resource stocks. Measures of resources must be on the cardinal scale and interpersonally comparable such that we can establish whether any two individuals hold unequal resource amounts.

How is this principle justified? All members of a society jointly decide on the fair distribution of goods and burdens. Given the moral equality of all members, it is immoral to enforce a distribution that is not accepted by all members. Enforcing the distribution requires that it is impartially justified \citep{gosepath_equality_2011}: No member asks for more from others than she is willing to give herself (reciprocity). All members have good reasons for accepting the distribution and no member has good reason for rejecting the distribution (universality). Any member who claims a larger share of the good has to provide an adequate reciprocal and universal justification for the claim. Unless there is an adequate justification for an unequal distribution, an equal distribution is the only legitimate distribution.

For many cases, strict egalitarianism is rejected on multiple grounds \citep{arneson_egalitarianism_2013, gosepath_equality_2011,lamont_distributive_2017}: (1) Equality undermines incentives to engage in wealth-creating activities because everyone gets the same irrespective of individual merit or contribution. In the long run, the stock of goods diminishes, making all individuals worse-off. The difference principle \citep{rawls_theory_2009} addresses this critique: Productive individuals are allowed to receive more resources but only to the extent that doing so will maximally improve the situation of the least-advantaged individuals. Inequality may be beneficial because it motivates the talented to make full use of their talent and, thereby, increases the stock of goods that society can distribute \citep{olsaretti_rawls_2018,wenar_john_2017}. (2) Equality neglects individuals’ responsibility for their actions and the ensuing consequences. A society bent on maintaining equality would require continuous transfers to individuals who squander their resources. (3) Equality is subject to the levelling down objection. Imagine allocations \((A,B)\), where \(A\) is the resource amount allocated to individual A and \(B\) is the resource amount allocated to individual B. If presented with two allocations \((1,10)\) and \((1,1)\), egalitarians would choose the second allocation because it is more equal, even if no individual is better-off in this allocation.

\subsection{Desert} \label{section32}
The distribution principle of desert-based justice requires the allocation of amount \(R\) of the resource to individual \(X\) iff \(X\) deserves \(R\). \(X\) deserves \(R\) by virtue of having attribute (desert-base) \(Y\) \citep{feldman_desert_2020,olsaretti_desert-based_2018}. Desert-bases are properties of an individual due to which the individual deserves resources. Desert-bases should be morally-relevant, i.e., we can evaluate them as good or bad (valuation condition). Only then can we say that an individual deserves more (less) resources because she has a property or performed an action that is deemed good (bad). We may further constrain desert-bases to properties for which individuals can be held responsible (responsibility condition). The responsibility condition is contented, an issue we return to below. Justice is realized if the absolute difference between what individuals deserve and what individuals actually have is minimized \citep[for a much more elaborate argument]{kagan_geometry_2012}. Measures of resources and deservingness must be on the cardinal scale and interpersonally comparable.

In situations concerned with the distribution of gains produced via cooperation (e.g., via production in the market), three broad desert-bases are often proposed: (a) the contribution of the individual to the production, (b) the effort expended by individuals in the production process, and (c) the costs or sacrifices that the individual incurred due to its production activity \citep{lamont_distributive_2017}. The higher the contribution, effort or sacrifice of an individual, the higher the share of the collective gain that the individual deserves. In a given application context, it is necessary to further specify and measure what counts as a contribution, effort or sacrifice.

How can we justify the desert principle? Starting from the moral equality of humans (Section \ref{section31}), we can argue that, out of respect, all humans should be treated as responsible beings. Humans should be viewed as morally responsible for their actions. Treating people as they deserve is a way of treating them as responsible beings \citep{olsaretti_desert-based_2018}: When we treat an individual as she deserves, we assume that the individual is deserving because she is responsible for her deservingness. This argument is limited in the sense that there are other ways of treating humans as responsible beings without treating them according to desert. Desert is sufficient but not necessary to treat humans with respect. Nonetheless, by appealing to responsibility, desert avoids some criticism that is advanced against egalitarianism. Additional justifications of the desert principle highlight its instrumental value \citep{olsaretti_desert-based_2018}: Desert encourages moral behaviour via rewards and discourages immoral behaviour via punishments. Furthermore, giving people what they deserve might be non-instrumentally and intrinsically good. Most would agree that a situation where the virtuous are rewarded and the vicious are penalized is better than a situation where the virtuous are penalized and the vicious are rewarded. However, our evaluative attitudes towards different allocation situations (even if widely shared) might be too subjective to serve as a stable justification of desert-based justice \citep{feldman_desert_2020}.

Desert-based justice has been criticized on multiple grounds: (1) Basing desert on responsibility is problematic because individuals can deserve things for which they are not responsible \citep{feldman_desert_2020}. It may be true that a hard-working student deserves a good grade because she is responsible for her hard work. But it might also be true that a customer deserves compensation from a retailer who unwittingly sold her a broken product, even if the customer is not responsible for being sold a broken product. More broadly, individuals cannot be held responsible for desert-bases that are due to brute luck (Section \ref{section36}). Suppose wages are allocated according to workers’ productive contribution. The productive contribution can be assumed to depend on workers’ natural talent, a factor most would consider the result of brute luck. Whenever separation of the luck and non-luck factors underlying desert-bases is not possible, it becomes difficult to find reasonable desert-bases. (2) In pluralistic societies, citizens will have different views about what constitutes a reasonable desert-base \citep{olsaretti_desert-based_2018}. It might be objectionably paternalist for a state to enforce a controversial conception of desert. This argument does not deny that there might be uncontroversial desert-bases. Moreover, some desert bases might be objectively reasonable for states to enforce, even if some citizens disagree.

\subsection{Sufficiency and Need} \label{section33}
The distribution principle of sufficiency demands that all individuals receive the amount of goods required to meet their basic needs \citep{olsaretti_sufficiency_2018}. Basic needs are said to be met if the allocated resource amount surpasses a given threshold. More formally: Allocate amount \(R\) of the resource to individual \(X\) iff \(X\) needs \(R\). \(X\) needs \(R\) iff the stock \(Y\) of resources that \(X\) has in the absence of \(R\) is below the threshold of resources that individuals should minimally have. Measures of resources must be on the cardinal scale.

Where egalitarianism claims that it is important that all have the same in relative terms (comparative perspective), sufficiency claims that it is important that all have enough in absolute terms (non-comparative perspective). For sufficiency, inequalities between individuals do not matter as long as the basic needs of all individuals are met. We do not care about inequalities between millionaires and billionaires. We care about inequalities between those below the threshold who do not have enough to meet their basic needs and those above the threshold who have more than enough.

What are the relevant basic needs? Basic needs refer to goods that are necessary or indispensable for human functioning \citep{olsaretti_sufficiency_2018}. Leading a recognizably human life is only possible if basic needs are met. \cite{doyal_theory_1991} suggest two basic needs: physical health and autonomy, i.e., the ability to deliberate and make decisions. Access to a range of intermediate resources is required to satisfy these basic needs. Intermediate resources include: nutritious food and clean water, protective housing, a non-hazardous work environment, a non-hazardous physical environment, appropriate healthcare, security in childhood, significant primary relationships, physical security, economic security, appropriate education, safe birth control, and safe child-bearing. This list provides a good starting point but should not be seen as final or exhaustive. More extensive demands are made by theories of relational equality \citep{anderson_what_1999}: All individuals should be furnished with the amount of resources that is sufficient for individuals to participate in society as equals free from coercion and domination. Inequalities are allowed as long as everyone can participate in society to a sufficient extent.

How can we justify sufficiency? The moral equality of humans can be taken to imply that all humans should be able to lead a recognizably human life, i.e., a life where the basic needs for human functioning (and relational equality) are met. In this view, moral equality does not require that everyone has the same but that everyone has enough. Additional justifications of the sufficiency principle highlight its instrumental value \citep{olsaretti_sufficiency_2018}: Human functioning is a precondition for individuals to participate in value-creating cooperation. To reap the benefits of cooperation, we should meet individuals’ basic needs and enable them to participate in cooperation.

Sufficiency has been criticized on multiple grounds \citep{olsaretti_sufficiency_2018}: (1) Picking the threshold that indicates what amount of resources is enough to meet basic needs is somewhat arbitrary. (2) Sufficiency offers no guidance on how to deal with inequality above the threshold. Sufficiency implies transfers of resources from individuals above the threshold to individuals below the threshold. Who should bear how much of the burdens associated with these transfers? Should relatively poor individuals just above the threshold contribute as much as super-rich individuals far above the threshold? In fact, sufficiency assigns no weight to the resources of individuals above the threshold. Even the smallest gain below the threshold would outweigh even the biggest gain above the threshold. (3) In societies where the basic needs of all individuals are met, sufficiency offers little guidance. In reaction, we might expand the scope of sufficiency to non-basic needs. This would require us to enforce (potentially controversial) conceptions of what the relevant non-basic needs are. (4) Sufficiency neglects the responsibility of individuals: In its most basic form, sufficiency would require us to continuously re-supply resources to individuals who carelessly squander their resources and, thereby, remain below the threshold.

\subsection{Prioritarianism} \label{section34}
The distribution principle of priority-based justice holds that allocating resources to individuals is morally more important the worse-off these individuals are in the absence of the additional resources \citep{parfit_equality_1997}. More formally: Allocate amount \(R\) of the resource to individual \(X\) iff the allocation of \(R\) to \(X\) has the highest moral value \(Y\) among all possible allocations. Prioritarianism assumes decreasing marginal moral value: The more resources an individual already has, the lower the moral value of allocating additional resources to the individual. Prioritarianism is non-comparative: The moral value of allocating resources to an individual depends only on the stock of resources that the individual already has, not on the stock of resources possessed by other individuals \citep{holtug_prioritarianism_2007}. Prioritarianism recommends that, if we are in the position to allocate a fixed amount of resources, we should allocate these resources to the worst-off individuals. Doing so will maximize the aggregate moral value of the allocation. As before, measures of resources must be on the cardinal scale.

How can we justify prioritarianism? As long as the worst-off individuals are in conditions below the threshold of a recognizably human life, prioritarianism recommends allocating resources to these worst-off individuals. In this respect, prioritarianism shares its justification with sufficiency. Proponents of prioritarianism further argue that prioritarianism is superior to egalitarianism and sufficiency \citep{adler_prioritarianism_2019,holtug_prioritarianism_2007}. In contrast to egalitarianism, prioritarianism is not subject to the levelling down objection. Prioritarians prefer the allocation that maximizes the aggregate moral value. If presented with the two allocations (1,10) and (1,1), prioritarians select the first allocation (remember: egalitarians select the second allocation that minimizes inequality). With respect to sufficiency, it is argued that inequalities above the threshold matter and that prioritarianism handles these inequalities well (at least better than egalitarianism). Even if all individuals are above the threshold, we might still find it unjust that some have much more than others. Hence, if we are in a position to (re)allocate resources, we should allocate them to the worst-off individuals.

How has prioritarianism been criticized? (1) Like strict egalitarianism and sufficiency, prioritarianism neglects the responsibility of individuals: Prioritarianism would require us to continuously re-supply resources to the worst-off individuals even if these individuals carelessly squander their resources and, thereby, remain the worst-off. (2) Maximization of the aggregate moral value is problematic. The aggregate moral value of assigning small amounts of resources to a large number of well-off individuals can be higher than the aggregate moral value of assigning a large amount of resources to one badly-off individual (numbers-win argument).

\subsection{Equality of Opportunity} \label{section35}
Equality of opportunity operates in situations where access to resources (or other benefits) is tied to positions and posts \citep{arneson_equality_2015}. Formal equality of opportunity requires that access to such positions is open to all applicants. The applicant best qualified for the position is selected. Qualification is defined as the ability to successfully pursue the goal associated with the position. Attributes of the applicants that are not associated with qualification should not influence the selection. One can object that the ability to obtain the required qualification depends on factors for which individuals are not responsible. For instance, the family in which the individual is born might lack the means to pay for the required qualification.

Fair equality of opportunity \citep{rawls_theory_2009} holds that individuals with the same native talent and willingness to productively use the talent should have equal opportunity to get the position. All individuals should be able to develop their talents to the fullest extent. Realizing fair equality of opportunity necessitates massive redistributions to offset the unequal starting conditions provided by the family of origin.

Abstracting from the two definitions, we specify the distribution principle of equality of opportunity as: Allocate equal opportunity to receive amount \(R\) of the resource to all individuals \(X\) with the same attribute \(Y\). Under fair equality of opportunity, we have to make sure that the opportunity to obtain attribute \(Y\) does not depend on morally arbitrary factors, such as sex, ethnicity, and family background.

\subsection{Luck and Responsibility} \label{section36}
The distribution principles of egalitarianism, sufficientarianism, and prioritarianism are criticized for neglecting the responsibility of individuals. What does it mean for an individual to be responsible for her condition? When can we reasonably say that an individual is not responsible because her condition is due to luck? What kinds of luck require redress?

Luck egalitarianism holds that unequal resource distributions resulting from luck are unjust and should be corrected \citep{olsaretti_dworkin_2018,lippert-rasmussen_justice_2018}. Here, we distinguish between option luck and brute luck (\citealt{dworkin_what_1981}; \citeyear{dworkin_what_1981b}). Inequality arising from brute luck should be equalized, inequality arising from option luck not. Option luck is the outcome of a deliberate gamble that the person could have anticipated and might have declined. Brute luck is the outcome of a gamble that is not deliberate because the person lacked a reasonable alternative to accepting the gamble (e.g., the natural lotteries of talent, family background, ethnicity) \citep{olsaretti_dworkin_2018,dworkin_what_1981b}. It has been suggested that bad option luck should be equalized too if the gamble is morally required or socially beneficial \citep{lippert-rasmussen_justice_2018}. For instance, a person might deliberately choose to stay at home to care for a child, risking that lifetime earnings will decrease. As childcare is socially beneficial, the person should be compensated in case of the bad option luck that lifetime earnings decreased. Even then, critics argue that luck egalitarianism is too unforgiving \citep{olsaretti_dworkin_2018}: People who experience bad luck often have a moral right to be helped, even if the bad luck came about via their own fault and choice.

In addition to option luck and brute luck, one can distinguish responsibility luck and desert luck \citep{lippert-rasmussen_justice_2018}. Responsibility luck states that \(Z\) (e.g., the outcome of a financial investment) is a matter of luck for \(X\) (a person) iff \(X\) is not morally responsible for \(Z\). Different versions of responsibility are possible. (a) Control responsibility: \(X\) is not responsible for \(Z\) iff \(X\) does not and did not control \(Z\). (b) Choice responsibility: \(X\) is not responsible for \(Z\) iff \(Z\) is not the result of a choice made by \(X\). One might say that the person is not control-responsible for the fortunate outcome of her financial investment (because she did not control whether the firm she invested in would be profitable) but choice-responsible (because she would not have enjoyed the fortunate outcome if she had not chosen to invest). In the first case, the fortunate outcome is considered luck that requires correction. In the second case, the fortunate outcome is not considered luck.

Desert luck states that \(Z\) is a matter of luck for \(X\) iff it is not the case that \(X\) deserves \(Z\). One may define desert such that \(X\) deserves \(Z\) iff it is fitting that \(X\) has \(Y\) given the moral merits of \(X\). For example, the person deserves the fortunate outcome of her investment because she is a talented and hard-working investor. In this case, the fortunate outcome is not considered luck and requires no correction.

To address the criticism advanced against egalitarianism, sufficientarianism, and prioritarianism, we can say that re-supplying resources to individuals who failed to benefit from their initial resource allocation is only justified if these individuals are not responsible for the failure.

Having reviewed various distributive justice theories, we next turn to fairness metrics commonly used in the fairML literature and discuss how to connect them to distributive justice.

\section{Fairness Metrics and Distributive Justice} \label{section4}
When presenting the setup and notation in Section 2, we carefully separated the prediction task from the decision task. The prediction task is handled by the prediction function \(g:V \rightarrow Y\). The decision task is governed by the distribution rule. The two tasks are related: A wrong prediction leads to a wrong decision. Nonetheless, the two tasks are associated with different distributive justice concerns: For the prediction task, we ask whether the distribution of prediction errors (defined below) is just. For the decision task, we ask whether the implemented distribution of resources is just. The distinction between prediction and decision is also evident in fairness metrics that aim to formalize different notions of distributive justice. This section shortly reviews existing fairness metrics. More extensive reviews are found in \cite{makhlouf_applicability_2020}, \cite{mitchell_prediction-based_2020}, and \cite{verma_fairness_2018}. Sections \ref{section41} and \ref{section42} trace the link between fairness metrics and distributive justice theories for the prediction task and the decision task, respectively.

Fairness in prediction-based decision-making concerns the question whether the output of prediction function \(g:V \rightarrow Y\) conforms to a given notion of justice. For the prediction task, prediction metrics check whether the probability of prediction errors differs across groups defined by protected attributes \(A\) like sex, ethnicity, and disability. Prediction errors are detected in a separate validation dataset of \(J\) individuals \(\{(v_1,y_1 ),…,(v_J,y_J)\}\) by comparing predicted target value \(g(v_i) = \hat{y}_i\) to true target value \(y_i\). For simplicity, assume a binary target such that \(y_i\) and \(\hat{y}_i \in \{0,1\}\). Predictions are correct if \(y_i = \hat{y}_i\), otherwise a prediction error occurred. Table \ref{tab2} presents popular prediction metrics.

\begin{table}
\caption{Definition and Illustration of Prediction Metrics}
\includegraphics[width=\textwidth]{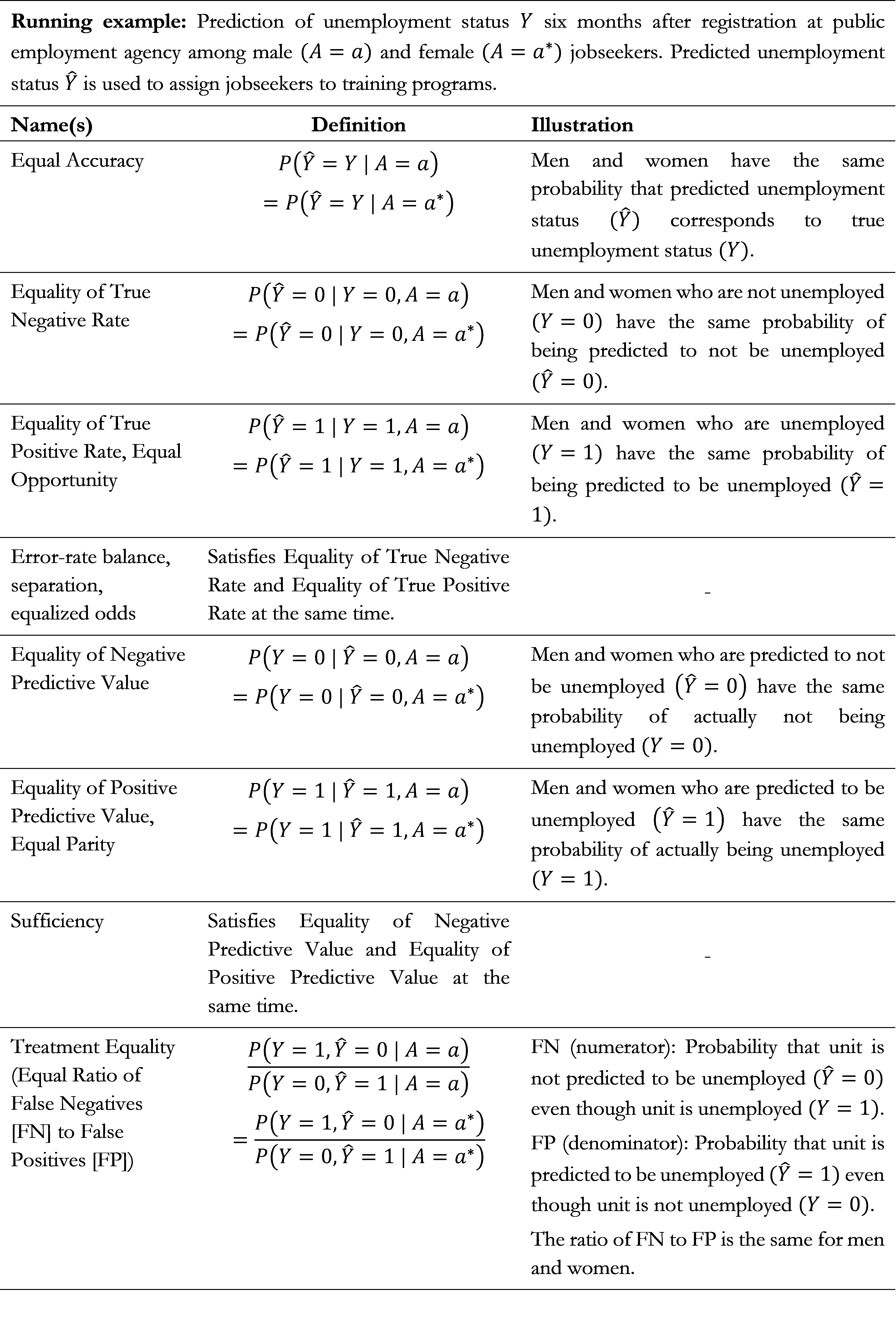}
\centering
\label{tab2}
\end{table}

For the decision task, decision metrics check whether the probability of being predicted into the positive class \((\hat{Y}=1)\) differs across groups defined by protected attributes \(A\) and, potentially, by unprotected attributes \(B\). Table \ref{tab3} presents decision metrics accompanied by illustrative examples.

\begin{table}
\caption{Definition and Illustration of Decision Metrics}
\includegraphics[width=\textwidth]{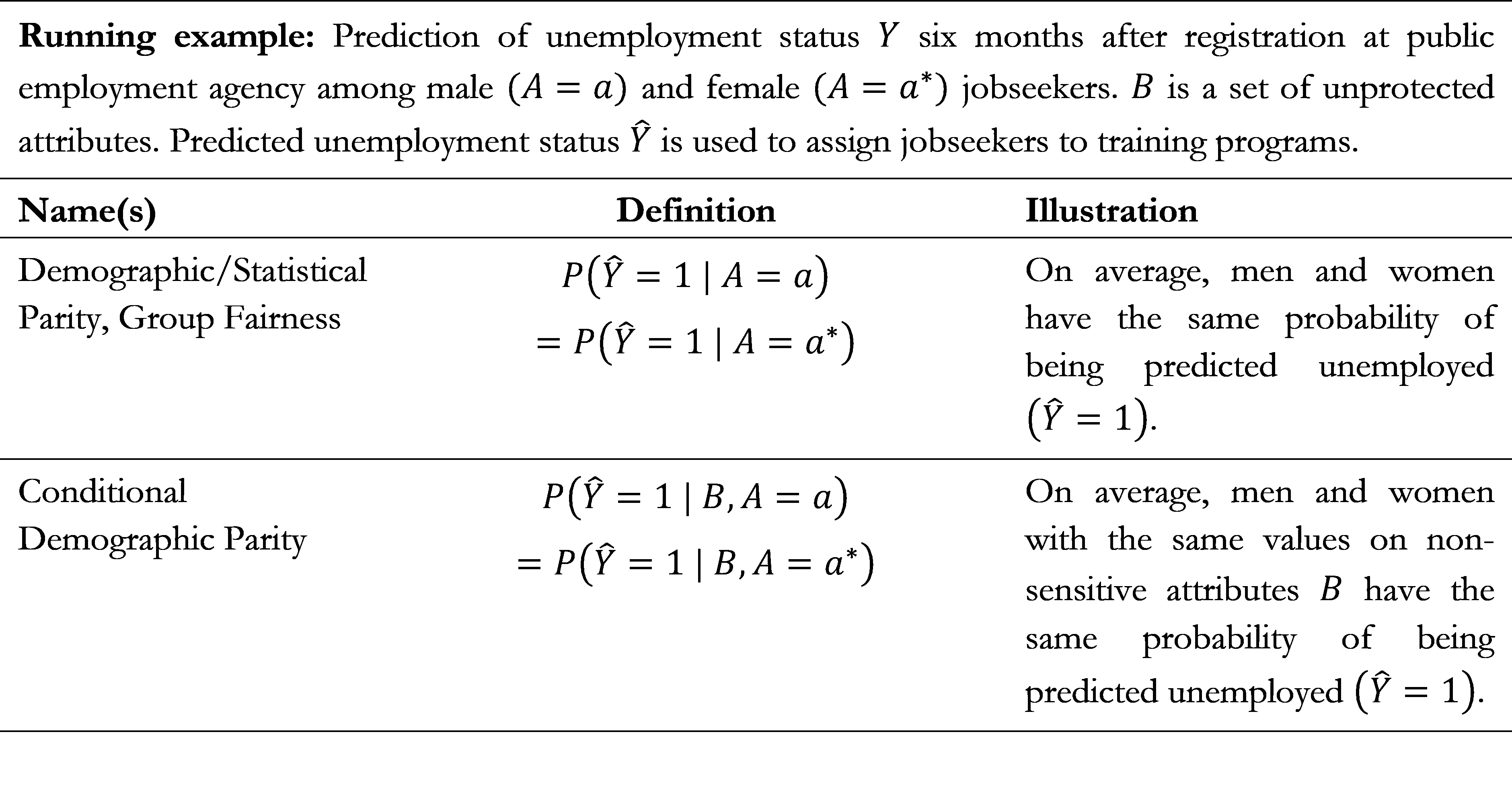}
\centering
\label{tab3}
\end{table}

\subsection{Distributive Justice and Prediction Metrics} \label{section41}
What is a justified rule for the distribution of prediction errors? Among the distribution rules discussed in Section \ref{section2}, we prefer strict egalitarianism: We should allocate the same probability of prediction error to all individuals, irrespective of any attribute \(Y\). Our reasoning is as follows: Individuals may deserve or need unequal amounts of resources (based on desert, sufficiency or priority). But all individuals should have an equal probability that the amount of resources that they should receive is determined without error. There is no obvious reason why some individuals should deserve or need a higher probability of prediction error than others. Which justice notion do prediction metrics (Table \ref{tab2}) capture?

Equality of True Negative Rate, Equality of True Positive Rate, and Error-Rate Balance map onto equality of opportunity (Section \ref{section35}). Individuals with the same qualification (same underlying true value \(Y\)) should have equal opportunities to realize a given outcome (equal chance of error-free prediction). Opportunities should not be affected by protected attributes like sex, ethnicity, and ability status that are morally arbitrary in the sense that the individual cannot be held responsible for having these attributes.

However, according to these metrics, the probability of prediction error is allowed to be unequal across groups defined by different underlying true values on \(Y\). In our running example (Box \ref{box:jobs}): Men and women who are not unemployed in six months \((Y=0)\) have the same probability of being predicted to not be unemployed \((\hat{Y}=0)\) (Equality of True Negative Rate). Men and women who are unemployed in six months \((Y=1)\) have the same probability of being predicted to be unemployed \((\hat{Y}=1)\) (Equality of True Positive Rate). But: The probability of correct prediction among those who will not be unemployed \((Y=0)\) can differ from the probability of correct prediction among those who will be unemployed \((Y=1)\). For example, erroneous decisions might be more frequent among those who will be unemployed. This result is rather unintuitive. Why should individuals with a higher value of \(Y\) deserve or need a different probability of prediction error than individuals with a lower value of \(Y\)? One might argue that \(Y\) captures some notion of qualification or merit that would entitle individuals to fewer prediction errors. However, the role of \(Y\) is not to capture entitlement to fewer prediction errors but entitlement to resources. The application of equal opportunity to the distribution of prediction errors seems misguided because it is difficult to identify attributes that could justify entitlement to fewer prediction errors. Equality of opportunity, a distribution rule designed to regulate the access to beneficial positions in stratified societies, does not map neatly onto the task of allocating prediction errors in automated allocation systems.

Equality of Negative Predictive Value, Equality of Positive Predictive Value, and Sufficiency (combination of the first two metrics) can also be mapped onto equality of opportunity. These metrics require that individuals who were assigned the same prediction \(\hat{Y}\) should have equal probabilities to have the same underlying true value \(Y\), irrespective of protected attributes for which individuals cannot be held responsible. For all practical purposes, these metrics can be viewed as equivalents to Equality of True Negative Rate, Equality of True Positive Rate, and Error-Rate Balance discussed above. As such, they face the same criticism.

Equal accuracy and Treatment Equality are better at capturing our egalitarian concern. The metrics require equal prediction errors across groups with different values on \(Y\). Further, the metrics demand equal probability of correct prediction (Equal Accuracy) or equal ratio of False Positives to False Negatives (Treatment Equality) across groups defined by protected attributes for which individuals cannot be held responsible. We object that prediction errors should also be equal across groups defined by unprotected attributes for which individuals can be held responsible. For the sake of the argument, assume that education is such an unprotected attribute. Why should individuals with different levels of education face different risks of prediction errors? Again, individuals with different levels of education may need or deserve different resource amounts. But the probability that the resource amount deserved or needed by these individuals is correctly predicted should be equal across educational levels. The same argument can be made for any unprotected attribute. Note that this perspective aligns with more recent, multi-group notions of fairness: Rather than focusing on a limited set of protected attributes, the notions of \cite{hebert-johnson2018} and \cite{kim2019} aim to improve the utility of a prediction model for arbitrary large sets of subpopulations defined by protected and/or unprotected attributes.

From the perspective of distributive justice, common prediction metrics lack a reasonable justification. Going back to the basic egalitarian argument (Section \ref{section31}): Unless there is an adequate justification for an unequal distribution, an equal distribution is the only legitimate distribution. In our view, no adequate justification for an unequal distribution of prediction errors can be made. Accordingly, showing that an automated allocation system satisfies existing prediction metrics is neither necessary nor sufficient for showing that the system is fair in any justified sense of distributive justice.

\subsection{Distributive Justice and Decision Metrics} \label{section42}
Which justice notion do decision metrics (Table \ref{tab3}) capture? Decision metrics aim to realize justice by guiding the selection of prediction models. Demographic Parity (DP) recommends prediction models where members of protected groups \(A\) and \(A^*\) have the same probability of receiving the positive prediction \((\hat{Y}=1)\). Conditional Demographic Parity (CDP) recommends prediction models where members of protected groups \(A\) and \(A^*\) with the same values on unprotected attributes \(B\) have the same probability of receiving the positive prediction \((\hat{Y}=1)\). The predicted attribute \(\hat{Y}\) is then plugged into the distribution rule to determine the resource allocation.

DP and CDP purposefully ignore true differences in the distribution of the prediction target \(Y\) across protected groups. Imagine that, due to gender discrimination on the labour market, female jobseekers have a higher probability than male jobseekers to remain unemployed \((Y=1)\) six months after registering at the public employment agency. Furthermore, assume that the true difference in re-employment probabilities also applies to male and female jobseekers with identical unprotected attributes like experience, education, and family status. DP and CDP recommend a prediction model where male and female jobseekers (with identical unprotected attributes) have the same probability of being predicted to remain unemployed \((\hat{Y}=1)\). The distribution rule allocates training programs to jobseekers based on predicted unemployment status \(\hat{Y}\). Hence, male and female jobseekers have the same probability of receiving training programs, irrespective of whether male and female jobseekers really face the same risk of remaining unemployed.

DP implements the notion that differences in the true prediction target \(Y\) between groups of individuals that are caused by protected attributes \(A\) should not influence the allocation of resources. CDP implements the notion that differences in the true prediction target \(Y\) between groups of individuals (with identical unprotected attributes \(B\)) that are caused by protected attributes \(A\) should not influence the allocation of resources. Through the lens of luck egalitarianism (Section \ref{section36}), protected attributes \(A\) are interpreted as attributes that are due to luck and for which individuals cannot be held responsible. Unprotected attributes \(B\) are interpreted as attributes for which individuals can be held responsible.

The luck egalitarian notion implemented by DP and CDP is justified in allocation scenarios where the responsibility of individuals is morally relevant. Concerns for responsibility are prevalent in desert-based distribution rules of the form: Allocate amount \(R\) of the resource to individual \(X\) iff \(X\) deserves \(R\). \(X\) deserves \(R\) by virtue of having attribute (desert-base) \(Y\). Whether or not an individual has the relevant desert-base \(Y\) may be partly caused by luck. Then, the individual is not fully responsible for having the desert base. In such scenarios, DP and CDP help identifying prediction models that are insensitive to luck that is captured by protected attributes.

Insisting that allocations only depend on attributes for which individuals can be held responsible is problematic. Individuals can deserve resources exactly because they are not responsible for their situation. For instance, female jobseekers deserve more support from the public employment agency exactly because they are subjected to labour market discrimination, a factor for which female jobseekers are hardly responsible. Similarly, priority and sufficiency advocate allocating resources to the worst-off individuals and to individuals who need resources to satisfy basic needs. It is not morally relevant whether individuals are responsible for being the worst-off or for having unmet basic needs. The distribution rules insist that individuals with different true values \(Y\) should be treated differently, even if the differences in \(Y\) are caused by factors for which the individuals are not responsible. Accordingly, we should prefer a prediction model that correctly represents true differences in \(Y\) in the predictions \(\hat{Y}\). If female jobseekers really have a lower probability of re-employment and, therefore, need more support, the prediction model should reflect this gender difference. Only then can the distribution rule allocate more support to female jobseekers.

Taken together, DP and CDP capture luck egalitarian notions of distributive justice. In allocation scenarios where the responsibility of individuals is morally relevant, DP and CDP offer useful guidance for the selection of prediction models. In scenarios where responsibility is not morally relevant, DP and CDP have the disagreeable consequence of ignoring relevant inequalities between individuals that would justify unequal treatment. In general, showing that an automated allocation system satisfies DP or CDP is neither necessary nor sufficient for showing that the system adheres to a justified notion of distributive justice other than luck egalitarianism.

\subsection{Related Approaches}
Before we turn to the discussion, we briefly review existing attempts that map fairness metrics to distributive justice theories and compare them to our mapping. Despite small differences, we conclude that our mapping is not exceptional.

\cite{heidari_moral_2019} cast the metrics Demographic Parity, Error-rate Balance, Equal Accuracy, and Sufficiency as different versions of equality of opportunity. The authors define equality of opportunity such that it contains luck egalitarianism as a border case. \cite{gajane_formalizing_2018} similarly map Equality of True Positive Rate (sub-dimension of Error-rate Balance) to equality of opportunity but associate Demographic Parity with egalitarianism. \cite{lee_fairness_2020} identify Equality of True Positive Rate, Equality of True Negative Rate, and Error-Rate Balance with equality of opportunity. Demographic Parity is mapped to egalitarianism. Taken together, the authors agree that the prediction metrics correspond to equality of opportunity. It is contested whether Demographic Parity represents egalitarianism or equality of opportunity.

With respect to prediction metrics, our mapping is similar. The only difference is that we do not understand Equal Accuracy as an implementation of equality of opportunity. Equal Accuracy equalizes prediction errors over protected groups, irrespective of the true target value \(Y\) and, therefore, it would be wrong to say that Equal Accuracy assigns prediction errors based on some measure of merit or qualification as required by equality of opportunity. We agree with \cite{heidari_moral_2019} that Demographic Parity corresponds to luck egalitarianism. Demographic parity is not strict egalitarianism as predictions are not equalized across all individuals but only across selected groups.

\section{Discussion} \label{section5}
Which conclusions should we draw regarding the connection between fairness metrics and distributive justice? Put plainly: The connection between fairness metrics and distributive justice theories is weak. Only two connections were found: (1) Prediction metrics apply equality of opportunity \citep{arneson_equality_2015} to the distribution of prediction errors. (2) Decision metrics implement notions of luck egalitarianism \citep{olsaretti_dworkin_2018,lippert-rasmussen_justice_2018}. We argued that these connections are defect or limited. (1) Equality of opportunity is not appropriate for the distribution of prediction errors because it allows objectionable inequalities in the probability of receiving correct predictions. (2) Luck egalitarianism, as implemented by decision metrics, applies only in a limited subclass of allocation scenarios. In this subclass, individuals’ responsibility for their situation is morally important. However, it is problematic that the decision metrics ignore inequalities between individuals that would justify unequal treatment in allocation scenarios where responsibility is not the main concern. (3) Luck egalitarianism is only one among multiple distributive justice theories. Concerns for egalitarianism, desert, sufficientarianism, and prioritarianism are not captured by existing decision metrics.

Apparently, the fair machine learning literature has not taken full advantage of the rich and longstanding literature on distributive justice. We believe that the main explanation for this shortcoming is that the fairML literature often does not distinguish between prediction and decision. Little surprising, surveys of the fair machine literature \citep{makhlouf_applicability_2020,mitchell_prediction-based_2020,verma_fairness_2018} do not distinguish the two tasks or focus exclusively on the prediction task. From the perspective of distributive justice, however, the decision task is of prime importance. In the decision task, we define the distribution rule: Allocate amount \(R\) of the resource to individual \(X\) iff \(X\) has attribute \(Y\). That is, we define the target attribute \(Y\) and the function that relates attribute \(Y\) to the resource amount \(R\) that should be allocated to individual \(X\). The prediction task is only of secondary importance (or even superfluous if attribute \(Y\) is observable at decision time). The decision task pre-supposes that the prediction task is carried out successfully such that accurate predictions of \(Y\) are obtained.

\begin{figure}[h]
\includegraphics[scale=0.5]{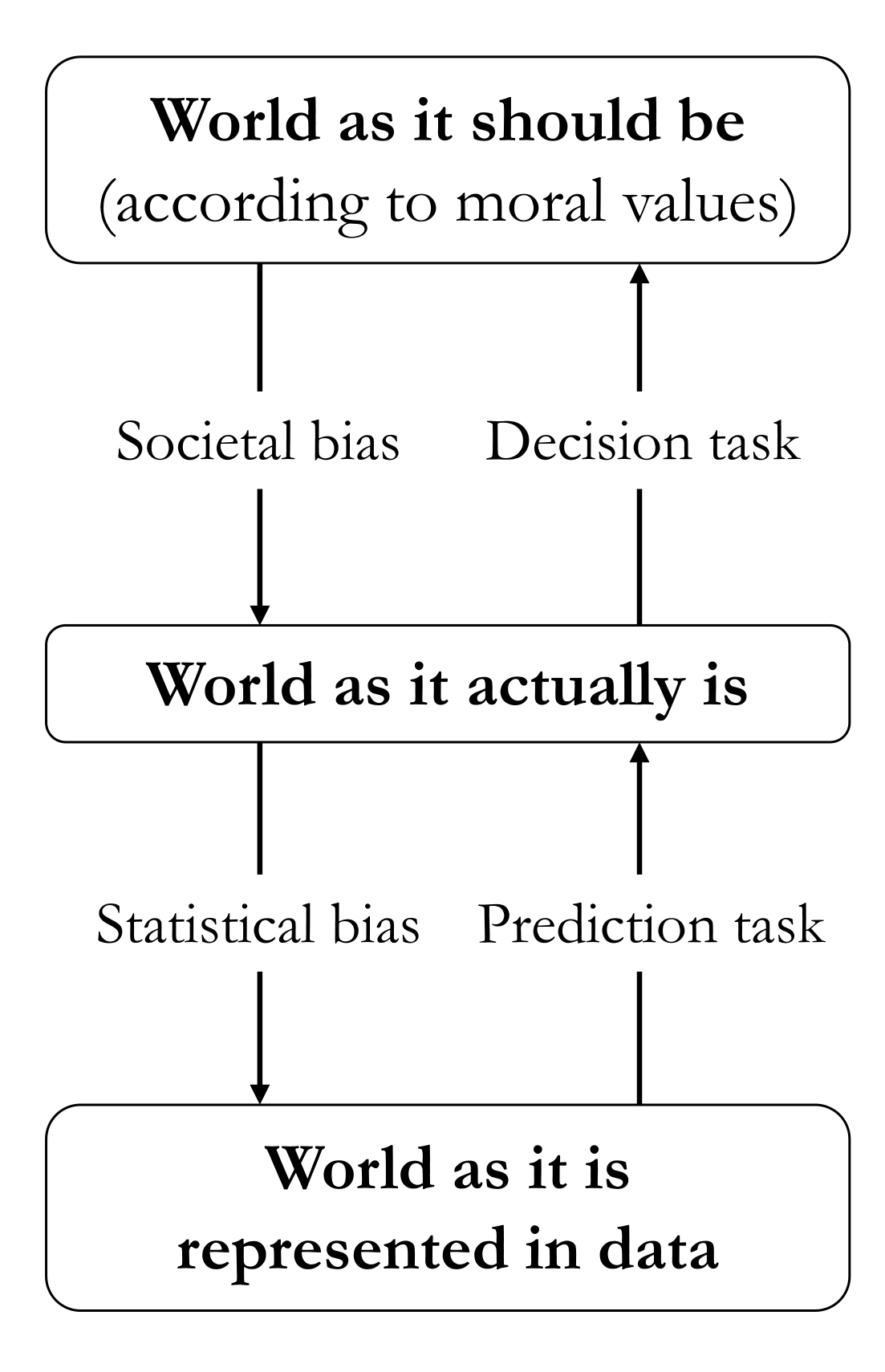}
\centering
\caption{Distinction Societal vs Statistical Bias and Decision vs Prediction Task}
\label{fig1}
\end{figure}

What are the appropriate roles of distributive justice theories and fairness metrics in automated allocation systems? To clarify the roles, it is useful to distinguish two types of biases that produce unintended allocation consequences: societal/historical bias and statistical bias \citep{mitchell_prediction-based_2020,suresh_framework_2020}. Societal bias, on the one hand. refers to the mismatch between how the world should be (according to our moral values, including, for example, distributive justice) and how the world actually is. Statistical bias, on the other hand, refers to the mismatch between how the world actually is and how the world is represented in the training and validation data. Statistical bias arises from representation/sampling bias (i.e., the data do not represent full population to which prediction model will be applied) and measurement bias (i.e., a systematic mismatch between measured attribute values and true attribute values). Figure \ref{fig1}, adapted from \cite{mitchell_prediction-based_2020}, illustrates the distinction between societal and statistical bias.

The prediction task's goal is to build an accurate model of how the world actually is from how the world is represented in the data. From a given training dataset \(\{(v_1,y_1 ),…,(v_N,y_N )\}\), we try to extract a rule \(g:V \rightarrow Y\) that accurately describes how attributes \(V\) are mapped to attribute \(Y\) in reality. Hence, for the prediction task, detecting and mitigating statistical bias is important. The prediction model is fair if it accurately captures how the world actually is for each individual.

The decision task's goal is to define a distribution rule that minimizes the distance between how the world actually is and how the world should be according to our moral values. The lion’s share of the moral work is done on the level of the decision task and not the prediction task. Correcting objectionable societal bias is the task of the distribution rules, not of the prediction model. The prediction model should accurately capture any existing objectionable societal bias such that a (hopefully) benevolent decision-maker can devise an appropriate distribution rule that corrects this bias.

Coming back to our running example: Imagine female jobseekers have lower re-employment chances than male jobseekers in reality. In the world as it should be according to egalitarianism, male and female jobseekers should have equal re-employment chances. Hence, there is social bias that needs correction (given that egalitarianism is justified in this application context). An appropriate distribution rule would allocate resources in such a way that re-employment chances are equalized across male and female jobseekers. Detecting social bias requires an accurate description of the world as it actually is such that it can be compared to the world as it should be. The only requirement for the prediction model is that it accurately represents the world as it actually is.

The role of fairness metrics is to detect societal bias, i.e., instances where the world as it actually is deviates from the world as it should be. It is not the role of fairness metrics to constrain model selection or optimization in ways such that the resulting prediction model no longer accurately represents the world as it actually is. The role of distributive justice is to define how the world should be. Once decision-makers defined a justified distribution rule, we need to formalize fairness metrics that detect deviations between the world as it actually is (represented by the prediction model) and the world as it should be (defined by distributive justice). Having found such deviations, decision-makers can define distribution rules that minimize the deviation. Given the richness of application contexts and available distributive justice theories, the focus on equalizing outcomes across a limited set of protected attributes seems short-sighted. Fairness metrics need to be more flexible if they are to function as the link between prediction models and distribution rules.

\section{Conclusion} \label{section6}
Our contribution to the literature is twofold: (1) We provided an accessible discussion of the most prominent distributive justice theories. (2) We systematically discussed the link between distributive justice and fairness metrics. Our main take-home message is that evaluations of fairness in automated decision-making should be accompanied by a careful distinction between the prediction task and the decision task. Conflating the two tasks resulted in an overly narrow conception of justice in the fair machine learning literature so far. Justice is not established by constraining the prediction function to produce equal output across protected groups. Justice is established by selecting an appropriate distribution rule and by making sure that the prediction function provides accurate predictions as input for the distribution rule.

There is no ideal and general answer to the question which distribution rule should be used in which context. Political philosophy has not converged on a single distributive justice theory (and probably never will) and weighty arguments have been advanced against all theories. Here, our conclusion is not that developers of automated allocation systems should ignore distributive justice. On the contrary, distinguishing between the prediction task and the decision task facilitates an open debate with all stakeholders involved on which allocation rules are most suitable and accepted for a given application. Moreover, those who are concerned with the technical aspects of the decision task should closely confer with those who have domain knowledge about the decision task. Even if no ideal and general solution is available, developers can eventually strive for the best non-ideal solution in the particular application context. Our review of the most prominent distributive justice theories, their justification, and their weaknesses prepares both stakeholders and developers for this task.

\bibliography{references}

\begin{thebibliography}{}

\bibitem[Adler and Holtug, 2019]{adler_prioritarianism_2019}
Adler, M.~D. and Holtug, N. (2019).
\newblock Prioritarianism: {A} response to critics.
\newblock {\em Politics, Philosophy \& Economics}, 18(2):101--144.

\bibitem[Allhutter et~al., 2020]{allhutter_algorithmic_2020}
Allhutter, D., Cech, F., Fischer, F., Grill, G., and Mager, A. (2020).
\newblock Algorithmic {Profiling} of {Job} {Seekers} in {Austria}: {How}
  {Austerity} {Politics} {Are} {Made} {Effective}.
\newblock {\em Frontiers in Big Data}, 3:5.

\bibitem[Anderson, 1999]{anderson_what_1999}
Anderson, E. (1999).
\newblock What {Is} the {Point} of {Equality}?
\newblock {\em Ethics}, 109(2):287--337.

\bibitem[Arneson, 2013]{arneson_egalitarianism_2013}
Arneson, R. (2013).
\newblock Egalitarianism.
\newblock In Zalta, E.~N., editor, {\em The {Stanford} {Encyclopedia} of
  {Philosophy}}. Metaphysics Research Lab, Stanford University, summer 2013
  edition.

\bibitem[Arneson, 2015]{arneson_equality_2015}
Arneson, R. (2015).
\newblock Equality of {Opportunity}.
\newblock In Zalta, E.~N., editor, {\em The {Stanford} {Encyclopedia} of
  {Philosophy}}. Metaphysics Research Lab, Stanford University, summer 2015
  edition.

\bibitem[Arneson, 2018]{olsaretti_dworkin_2018}
Arneson, R.~J. (2018).
\newblock {\em Dworkin and {Luck} {Egalitarianism}}, volume~1.
\newblock Oxford University Press.

\bibitem[Barocas and Selbst, 2016]{barocas_big_2016}
Barocas, S. and Selbst, A.~D. (2016).
\newblock Big {Data}'s {Disparate} {Impact}.
\newblock {\em California Law Review}, 104(3):671--732.
\newblock Publisher: California Law Review, Inc.

\bibitem[Binns, 2018]{binns_fairness_2018}
Binns, R. (2018).
\newblock Fairness in {Machine} {Learning}: {Lessons} from {Political}
  {Philosophy}.
\newblock {\em arXiv:1712.03586 [cs]}.
\newblock arXiv: 1712.03586.

\bibitem[Brock, 2018]{olsaretti_sufficiency_2018}
Brock, G. (2018).
\newblock {\em Sufficiency and {Needs}-{Based} {Approaches}}, volume~1.
\newblock Oxford University Press.

\bibitem[Doyal and Gough, 1991]{doyal_theory_1991}
Doyal, L. and Gough, I. (1991).
\newblock {\em A {Theory} of {Human} {Need}}.
\newblock Macmillan Education UK, London.

\bibitem[Dworkin, 1981a]{dworkin_what_1981}
Dworkin, R. (1981a).
\newblock What is {Equality}? {Part} 1: {Equality} of {Welfare}.
\newblock {\em Philosophy \& Public Affairs}, 10(3):185--246.
\newblock Publisher: Wiley.

\bibitem[Dworkin, 1981b]{dworkin_what_1981b}
Dworkin, R. (1981b).
\newblock What is {Equality}? {Part} 2: {Equality} of {Resources}.
\newblock {\em Philosophy \& Public Affairs}, 10(4):283--345.
\newblock Publisher: Wiley.

\bibitem[Feldman and Skow, 2020]{feldman_desert_2020}
Feldman, F. and Skow, B. (2020).
\newblock Desert.
\newblock In Zalta, E.~N., editor, {\em The {Stanford} {Encyclopedia} of
  {Philosophy}}. Metaphysics Research Lab, Stanford University, winter 2020
  edition.

\bibitem[Freeman, 2018]{olsaretti_rawls_2018}
Freeman, S. (2018).
\newblock {\em Rawls on {Distributive} {Justice} and the {Difference}
  {Principle}}, volume~1.
\newblock Oxford University Press.

\bibitem[Gajane and Pechenizkiy, 2018]{gajane_formalizing_2018}
Gajane, P. and Pechenizkiy, M. (2018).
\newblock On {Formalizing} {Fairness} in {Prediction} with {Machine}
  {Learning}.
\newblock {\em arXiv:1710.03184 [cs, stat]}.
\newblock arXiv: 1710.03184.

\bibitem[Gerstner, 2018]{gerstner_predictive_2018}
Gerstner, D. (2018).
\newblock Predictive {Policing} in the {Context} of {Residential} {Burglary}:
  {An} {Empirical} {Illustration} on the {Basis} of a {Pilot} {Project} in
  {Baden}-{Württemberg}, {Germany}.
\newblock {\em European Journal for Security Research}, 3(2):115--138.

\bibitem[Gosepath, 2011]{gosepath_equality_2011}
Gosepath, S. (2011).
\newblock Equality.
\newblock In Zalta, E.~N., editor, {\em The {Stanford} {Encyclopedia} of
  {Philosophy}}. Metaphysics Research Lab, Stanford University, spring 2011
  edition.

\bibitem[Hebert-Johnson et~al., 2018]{hebert-johnson2018}
Hebert-Johnson, U., Kim, M., Reingold, O., and Rothblum, G. (2018).
\newblock Multicalibration: Calibration for the
  ({C}omputationally-identifiable) masses.
\newblock In Dy, J. and Krause, A., editors, {\em Proceedings of the 35th
  International Conference on Machine Learning}, volume~80 of {\em Proceedings
  of Machine Learning Research}, pages 1939--1948, Stockholmsmässan, Stockholm
  Sweden. PMLR.

\bibitem[Heidari et~al., 2019]{heidari_moral_2019}
Heidari, H., Loi, M., Gummadi, K.~P., and Krause, A. (2019).
\newblock A {Moral} {Framework} for {Understanding} {Fair} {ML} through
  {Economic} {Models} of {Equality} of {Opportunity}.
\newblock In {\em Proceedings of the {Conference} on {Fairness},
  {Accountability}, and {Transparency}}, pages 181--190, Atlanta GA USA. ACM.

\bibitem[Holtug, 2007]{holtug_prioritarianism_2007}
Holtug, N. (2007).
\newblock Prioritarianism.
\newblock In Holtug, N. and Lippert-Rasmussen, K., editors, {\em
  Egalitarianism: new essays on the nature and value of equality}, pages
  125--156. Clarendon Press, Oxford ; New York.
\newblock OCLC: ocm71807832.

\bibitem[Howard and Dixon, 2012]{howard_construction_2012}
Howard, P.~D. and Dixon, L. (2012).
\newblock The {Construction} and {Validation} of the {OASys} {Violence}
  {Predictor}: {Advancing} {Violence} {Risk} {Assessment} in the {English} and
  {Welsh} {Correctional} {Services}.
\newblock {\em Criminal Justice and Behavior}, 39(3):287--307.

\bibitem[Kagan, 2012]{kagan_geometry_2012}
Kagan, S. (2012).
\newblock {\em The {Geometry} of {Desert}}.
\newblock Oxford University Press.

\bibitem[Kim et~al., 2019]{kim2019}
Kim, M.~P., Ghorbani, A., and Zou, J. (2019).
\newblock Multiaccuracy: Black-box post-processing for fairness in
  classification.
\newblock In {\em Proceedings of the 2019 AAAI/ACM Conference on AI, Ethics,
  and Society}, AIES '19, page 247–254, New York, NY, USA. Association for
  Computing Machinery.

\bibitem[Kymlicka, 2002]{kymlicka_contemporary_2002}
Kymlicka, W. (2002).
\newblock {\em Contemporary political philosophy: an introduction}.
\newblock Oxford University Press, Oxford ; New York, 2nd ed edition.

\bibitem[Lamont and Favor, 2017]{lamont_distributive_2017}
Lamont, J. and Favor, C. (2017).
\newblock Distributive {Justice}.
\newblock In Zalta, E.~N., editor, {\em The {Stanford} {Encyclopedia} of
  {Philosophy}}. Metaphysics Research Lab, Stanford University, winter 2017
  edition.

\bibitem[Lee et~al., 2020]{lee_fairness_2020}
Lee, M. S.~A., Floridi, L., and Singh, J. (2020).
\newblock From {Fairness} {Metrics} to {Key} {Ethics} {Indicators} ({KEIs}):
  {A} {Context}-{Aware} {Approach} to {Algorithmic} {Ethics} in an {Unequal}
  {Society}.
\newblock {\em SSRN Electronic Journal}.

\bibitem[Lepri et~al., 2018]{lepri_fair_2018}
Lepri, B., Oliver, N., Letouzé, E., Pentland, A., and Vinck, P. (2018).
\newblock Fair, {Transparent}, and {Accountable} {Algorithmic}
  {Decision}-making {Processes}: {The} {Premise}, the {Proposed} {Solutions},
  and the {Open} {Challenges}.
\newblock {\em Philosophy \& Technology}, 31(4):611--627.

\bibitem[Lippert-Rasmussen, 2018]{lippert-rasmussen_justice_2018}
Lippert-Rasmussen, K. (2018).
\newblock Justice and {Bad} {Luck}.
\newblock In Zalta, E.~N., editor, {\em The {Stanford} {Encyclopedia} of
  {Philosophy}}. Metaphysics Research Lab, Stanford University, summer 2018
  edition.

\bibitem[Makhlouf et~al., 2020]{makhlouf_applicability_2020}
Makhlouf, K., Zhioua, S., and Palamidessi, C. (2020).
\newblock On the {Applicability} of {ML} {Fairness} {Notions}.
\newblock {\em arXiv:2006.16745 [cs, stat]}.
\newblock arXiv: 2006.16745.

\bibitem[Mehrabi et~al., 2019]{mehrabi_survey_2019}
Mehrabi, N., Morstatter, F., Saxena, N., Lerman, K., and Galstyan, A. (2019).
\newblock A {Survey} on {Bias} and {Fairness} in {Machine} {Learning}.
\newblock {\em arXiv:1908.09635 [cs]}.
\newblock arXiv: 1908.09635.

\bibitem[Miller, 2017]{miller_justice_2017}
Miller, D. (2017).
\newblock Justice.
\newblock In Zalta, E.~N., editor, {\em The {Stanford} {Encyclopedia} of
  {Philosophy}}. Metaphysics Research Lab, Stanford University, fall 2017
  edition.

\bibitem[Mitchell et~al., 2020]{mitchell_prediction-based_2020}
Mitchell, S., Potash, E., Barocas, S., D'Amour, A., and Lum, K. (2020).
\newblock Prediction-{Based} {Decisions} and {Fairness}: {A} {Catalogue} of
  {Choices}, {Assumptions}, and {Definitions}.
\newblock {\em arXiv:1811.07867 [stat]}.
\newblock arXiv: 1811.07867.

\bibitem[Moriarty, 2018]{olsaretti_desert-based_2018}
Moriarty, J. (2018).
\newblock {\em Desert-{Based} {Justice}}, volume~1.
\newblock Oxford University Press.

\bibitem[Niklas et~al., 2015]{niklas_profiling_2015}
Niklas, J., Sztandar-Sztanderskal, K., and Szymielewicz, K. (2015).
\newblock Profiling the unemployed in {Poland}: Social and political
  implications of algorithmic decison making.
\newblock Technical report, Fundacja Panoptykon.

\bibitem[Olsaretti, 2018]{olsaretti_introduction_2018}
Olsaretti, S. (2018).
\newblock {\em Introduction}, volume~1.
\newblock Oxford University Press.

\bibitem[Parfit, 1997]{parfit_equality_1997}
Parfit, D. (1997).
\newblock Equality and {Priority}.
\newblock {\em Ratio}, 10(3):202--221.

\bibitem[Rawls, 2009]{rawls_theory_2009}
Rawls, J. (2009).
\newblock {\em A {Theory} of {Justice}}.
\newblock Harvard University Press.

\bibitem[Robeyns and Byskov, 2020]{robeyns_capability_2020}
Robeyns, I. and Byskov, M.~F. (2020).
\newblock The {Capability} {Approach}.
\newblock In Zalta, E.~N., editor, {\em The {Stanford} {Encyclopedia} of
  {Philosophy}}. Metaphysics Research Lab, Stanford University, winter 2020
  edition.

\bibitem[Rudin, 2019]{rudin_stop_2019}
Rudin, C. (2019).
\newblock Stop explaining black box machine learning models for high stakes
  decisions and use interpretable models instead.
\newblock {\em Nature Machine Intelligence}, 1(5):206--215.

\bibitem[Suresh and Guttag, 2020]{suresh_framework_2020}
Suresh, H. and Guttag, J.~V. (2020).
\newblock A {Framework} for {Understanding} {Unintended} {Consequences} of
  {Machine} {Learning}.
\newblock {\em arXiv:1901.10002 [cs, stat]}.
\newblock arXiv: 1901.10002.

\bibitem[Verma and Rubin, 2018]{verma_fairness_2018}
Verma, S. and Rubin, J. (2018).
\newblock Fairness definitions explained.
\newblock In {\em Proceedings of the {International} {Workshop} on {Software}
  {Fairness}}, pages 1--7, Gothenburg Sweden. ACM.

\bibitem[Walen, 2020]{walen_retributive_2020}
Walen, A. (2020).
\newblock Retributive {Justice}.
\newblock In Zalta, E.~N., editor, {\em The {Stanford} {Encyclopedia} of
  {Philosophy}}. Metaphysics Research Lab, Stanford University, fall 2020
  edition.

\bibitem[Wenar, 2017]{wenar_john_2017}
Wenar, L. (2017).
\newblock John {Rawls}.
\newblock In Zalta, E.~N., editor, {\em The {Stanford} {Encyclopedia} of
  {Philosophy}}. Metaphysics Research Lab, Stanford University, spring 2017
  edition.

\end{thebibliography}
\end{document}